\documentclass{article}
\usepackage{PRIMEarxiv}

\usepackage[utf8]{inputenc}
\usepackage[T1]{fontenc}
\usepackage{hyperref}
\usepackage{amsfonts}
\usepackage{graphicx}
\usepackage{hhline}         
\usepackage{fancyvrb}       
\usepackage{float}          
\usepackage{enumitem}
\usepackage{subcaption}
\setitemize{noitemsep,topsep=.5em,parsep=.5em,partopsep=.1em}

\usepackage{authblk}

\newcommand{\noaffil}{ \hspace{-.7ex}}

\newcommand\Tstrut{\rule{0pt}{2ex}}         

\usepackage[frozencache, cachedir=minted-output]{minted} 
\newlength{\mintednumbersep}
\AtBeginDocument{%
  \sbox0{\tiny00}%
  \setlength\mintednumbersep{\parindent}%
  \addtolength\mintednumbersep{-\wd0}%
}
\usepackage{xcolor}
\definecolor{LightGray}{gray}{0.95} 

\usepackage[]{biblatex}
\newcommand{\mod}{\ensuremath{\%}}

\title{A Configurable Library for Generating and Manipulating Maze Datasets
}

\author[*,1]{Michael Igorevich Ivanitskiy} 
\author[\noaffil]{Rusheb Shah} 
\author[2]{Alex F. Spies} 
\author[\noaffil]{Tilman Räuker} 
\author[\noaffil]{Dan Valentine} 
\author[\noaffil]{Can Rager} 
\author[\noaffil]{Lucia Quirke} 
\author[\noaffil]{Chris Mathwin} 
\author[\noaffil]{Guillaume Corlouer} 
\author[1]{Cecilia Diniz Behn} 
\author[1]{Samy Wu Fung} 

\affil[*]{Corresponding Author: \href{mailto:mivanits@umich.edu}{mivanits@umich.edu}}
\affil[1]{Colorado School of Mines, Department of Applied Mathematics and Statistics}
\affil[2]{Imperial College London}

\begin{document}

\maketitle

\begin{abstract}
Understanding how machine learning models respond to distributional shifts is a key research challenge. 
Mazes serve as an excellent testbed due to varied generation algorithms offering a nuanced platform to simulate both subtle and pronounced distributional shifts. 
To enable systematic investigations of model behavior on out-of-distribution data, we present \texttt{maze-dataset}, a comprehensive library for generating, processing, and visualizing datasets consisting of maze-solving tasks. 
With this library, researchers can easily create datasets, having extensive control over the generation algorithm used, the parameters fed to the algorithm of choice, and the filters that generated mazes must satisfy. 
Furthermore, it supports multiple output formats, including rasterized and text-based, catering to convolutional neural networks and autoregressive transformer models. 
These formats, along with tools for visualizing and converting between them, ensure versatility and adaptability in research applications.
\end{abstract}

\hypertarget{introduction}{%
\section{Introduction}\label{introduction}
}

Out-of-distribution generalization is a critical challenge in modern machine learning (ML) research. 
For interpretability and behavioral research in this area, training on algorithmic tasks offers benefits by allowing systematic data generation and task decomposition, as well as simplifying the process of circuit discovery~\cite{interpretability-survery}.
Although mazes are well suited for these investigations, we have found that existing maze generation packages~\cite{cobbe2019procgen,harriesMazeExplorerCustomisable3D2019,gh_Ehsan_2022,gh_Nemeth_2019,easy_to_hard} do not provide support in flexibility of maze generation algorithms with fine-grained control of generation parameters and the ability to easily transform between multiple representations of the mazes (Images, Textual, Tokenized) for training and testing models.

This work aims to facilitate deeper research into generalization and interpretability by addressing these limitations. 
We introduce \href{https://github.com/understanding-search/maze-dataset}{\inlinecode{maze-dataset}}, an accessible Python package~\cite{maze-dataset-github}. 
This package offers flexible configuration options for maze dataset generation, allowing users to select from a range of algorithms and adjust corresponding parameters (Section~\ref{generation}). 
Furthermore, it supports various output formats tailored to different ML architectures (Section~\ref{output-formats}). 

\begin{figure}[H]
    \centering
    \makebox[\textwidth][c]{\includegraphics[width=0.9\textwidth]{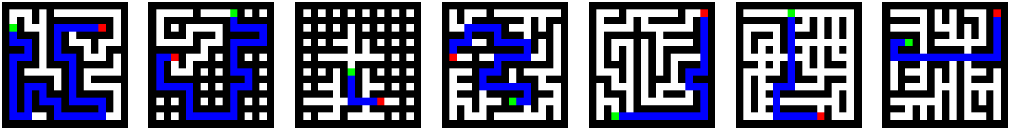}}%
    \caption{Example mazes from various algorithms. Left to right: randomized depth-first search (RDFS), RDFS without forks, constrained RDFS, Wilson's \cite{wilson}, RDFS with percolation ($p=0.1$), RDFS with percolation ($p=0.4$), random stack RDFS. Further examples available in the appendix of this work (Section~\ref{appendix}).}
    \label{fig:example_mazes}
\end{figure}

\newpage

\hypertarget{generation}{%
\section{Maze Generation and Usage}\label{generation}}

Our package can be installed from \href{https://pypi.org/project/maze-dataset/}{PyPi} via \texttt{pip install maze-dataset}, or directly from the \href{https://github.com/understanding-search/maze-dataset}{git repository} \cite{maze-dataset-github}.

To create a dataset, we first create a \texttt{MazeDatasetConfig} configuration object, which specifies the seed, number, and size of mazes, as well as the generation algorithm and its corresponding parameters. This object is passed to a \texttt{MazeDataset} class to create a dataset. Crucially, this \texttt{MazeDataset} inherits from a PyTorch \cite{pytorch} \texttt{Dataset}, and can thus be easily incorporated into existing data pre-processing and training pipelines, e.g., through the use of a \texttt{Dataloader} class.

\begin{minted}[
    % xleftmargin=\parindent,
    frame=lines,
    framesep=2mm,
    baselinestretch=1.2,
    bgcolor=LightGray,
    fontsize=\footnotesize,
    % linenos
]{python}
from maze_dataset import MazeDataset, MazeDatasetConfig, LatticeMazeGenerators
cfg: MazeDatasetConfig = MazeDatasetConfig(
    name="example", 
    grid_n=3, 
    n_mazes=32, 
    maze_ctor=LatticeMazeGenerators.gen_dfs,
)
dataset: MazeDataset = MazeDataset.from_config(cfg)
\end{minted}

When initializing mazes, further configuration options can be specified through the \texttt{from\_config()} factory method as necessary. Options include 1) whether to generate the dataset during runtime or load an existing dataset, 2) if and how to parallelize generation, and 3) where to store the generated dataset. Full documentation of this is available in our repository \cite{maze-dataset-github}. Available maze generation algorithms are static methods of the \texttt{LatticeMazeGenerators} class and include the following:

\begin{itemize}
    \item
        \texttt{gen\_dfs} \textbf{(randomized depth-first search)}:
        Parameters can be passed to constrain the number of accessible cells, the number of forks in the maze, and the maximum tree depth. Creates a spanning tree by default or a partially spanning tree if constrained.
    \item
        \texttt{gen\_wilson} \textbf{(Wilson's algorithm)}: 
        Generates a random spanning tree via loop-erased random walk \cite{wilson}. 
    \item
        \texttt{gen\_percolation} \textbf{(percolation)}:
        Starting with no connections, every possible lattice connection is set to either true or false with some probability $p$, independently of all other connections. For the kinds of graphs that this process generates, we refer to existing work \cite{percolation, percolation-clustersize}. 
    \item
        \texttt{gen\_dfs\_percolation} \textbf{(randomized depth-first search with percolation)}: 
        A connection exists if it exists in a maze generated via \texttt{gen\_dfs OR gen\_percolation}. Useful for generating mazes that are not acyclic graphs.
    \end{itemize}

Furthermore, a dataset of mazes can be filtered to satisfy certain properties: 
\begin{minted}[
    % xleftmargin=\parindent,
    frame=lines,
    framesep=2mm,
    baselinestretch=1.2,
    bgcolor=LightGray,
    fontsize=\footnotesize,
    % linenos
]{python}
    dataset_filtered: MazeDataset = dataset.filter_by.path_length(min_length=3)
\end{minted}

Custom filters can be specified, and several filters are included:

\begin{itemize}
\item
  \texttt{path\_length(min\_length:\ int)}: shortest length from the
  origin to target should be at least \texttt{min\_length}.
\item
  \texttt{start\_end\_distance(min\_distance:\ int)}: Manhattan distance
  between start and end should be at least \texttt{min\_distance},
  ignoring walls.
\item
  \texttt{remove\_duplicates(...)}: remove mazes which are similar to
  others in the dataset, measured via Hamming distance.
\item
  \texttt{remove\_duplicates\_fast()}: remove mazes which are exactly
  identical to others in the dataset.
\end{itemize}

All implemented maze generation algorithms are stochastic by nature. For reproducibility, the \texttt{seed} parameter of \texttt{MazeDatasetConfig} may be set. In practice, we do not find that exact duplicates of mazes are generated with any meaningful frequency,
even when generating large datasets.

\newpage

\hypertarget{output-formats}{%
\section{Output Formats}\label{output-formats}}

Internally, mazes are \texttt{SolvedMaze} objects, which have path information, and a connection list optimized for storing sub-graphs of a lattice. These objects can be converted to and from several formats.

\begin{figure}[H]
    \centering
    \begin{tabular}{p{2in} p{2in} p{2in}}
        \hline \\[.5em]
        \texttt{as\_ascii()} & \texttt{as\_pixels()} & \texttt{MazePlot()} \\[.5em]
            Simple text format for displaying mazes, useful for debugging in a terminal environment.
            & \texttt{numpy} array of \texttt{dtype=uint8} and shape \texttt{(height, width, 3)}. The last dimension is RGB color.
            & feature-rich plotting utility with support for multiple paths, heatmaps over positions, and more. \\[1em]
        \hline \\
            \multicolumn{1}{c}{\begin{minipage}[b]{1.6in}
                    \input{figures/outputs-ascii-colored.tex} 
            \end{minipage}}
            & \multicolumn{1}{c}{
                \includegraphics[width=0.25\textwidth]{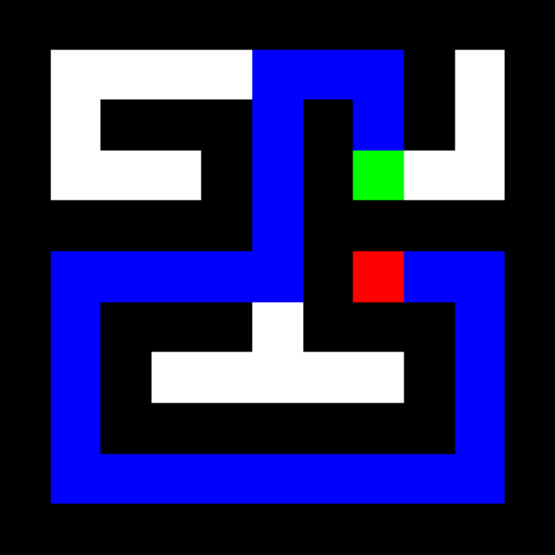}
            }
            & \multicolumn{1}{c}{
                \includegraphics[width=0.25\textwidth, trim={0 0 -.3cm, -.5cm}, clip]{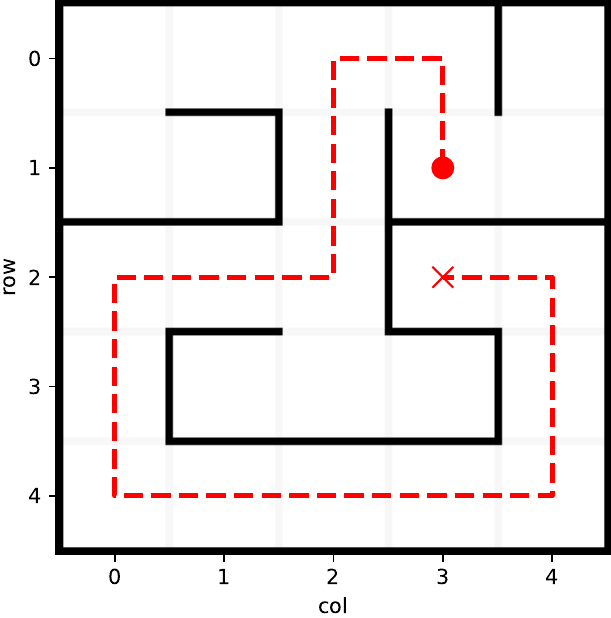}
            } \\[1em]
        \hline\hline \\
            \multicolumn{3}{l}{
                \texttt{as\_tokens()}
            } \\[.5em]
            \multicolumn{3}{p{6.3in}}{
                Text-based formats intended for autoregressive transformers. Several deterministic tokenization schemes (Section~\ref{training}) with scalable vocabularies are provided. The order of the adjacency list is, by default, randomized. Also included are utilities for colorizing the tokens according to the part of the task they represent. Colorizing can be done for HTML, \LaTeX, or terminal output.
            } \\[1em]
            \hline \\[.1em]
            \\[-1em]
            \multicolumn{3}{c}{
                \begin{minipage}{5.5in}
                    \fontsize{8}{1}
                    \input{figures/outputs-tokens-colored.tex}
                \end{minipage}
            } \\
        \hline \\[.5em]
    \end{tabular}
    \caption{Various output formats. Top row (left to right): ASCII diagram, rasterized pixel grid, and advanced display. Bottom row: text format for autoregressive networks.}
    \label{fig:output-fmts}
\end{figure}

\hypertarget{training}{%
\subsection{Training and Evaluation}\label{training}}

\begin{figure}[H]
    \centering
    \includegraphics[width=0.3\textwidth]{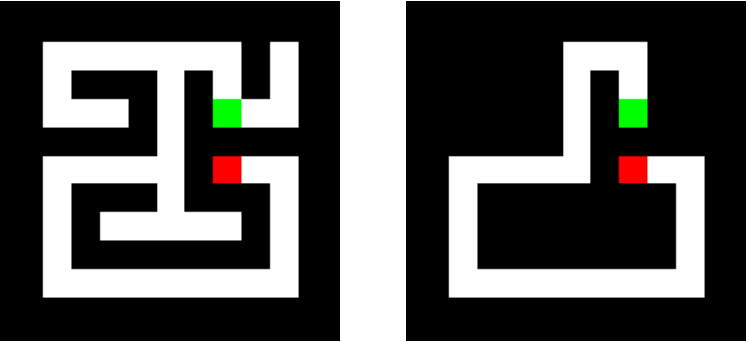}
    \caption{Input is the rasterized maze without the path marked (left), and provide as a target the maze with all but the correct path removed. Configuration options exist to adjust whether endpoints are included and if empty cells should be filled in.}
    \label{fig:output-rasterized}
\end{figure}

There are examples in the literature for training Recurrent Convolutional Neural Network (RCNN) derived architectures on maze tasks \cite{deepthinking}. To this end, we replicate the format of \cite{easy_to_hard} and provide the \texttt{RasterizedMazeDataset} class, which returns rasterized pairs of (input, target) mazes as shown in Figure \ref{fig:output-rasterized}.

To train autoregressive text models such as transformers, we use the full sequences provided by \texttt{as\_tokens()} shown in Figure \ref{fig:output-fmts}. During deployment we provide only the prompt up to the \texttt{<PATH\_START>} token. 
To map the vocabulary onto indices, we first allocate a portion of the indices for the ``special'' tokens which these do not represent coordinates. Next, we add each coordinate as a unique token. 
Coordinates are ordered in the vocabulary such that a maze of size $m$ will be processed the same way as the top $m \times m$ cells of a size-$n$ maze, where $n > m$. 
This is done so that models can be deployed on mazes smaller than the training size without destroying the structure of the vocabulary. 
Examples of usage of this dataset to train autoregressive transformers can be found in our \texttt{maze-transformer} library \cite{maze-transformer-github}. Other tokenization and vocabulary schemes are also included, such as representing each coordinate as a pair of $i,j$ index tokens.

\begin{figure}[H]
    \centering
    \begin{tabular}{p{4.3in} p{1.5in}}
        \begin{minipage}{4.3in}
            \fontsize{6}{1}
            \input{figures/prompt-tokens-colored}
        \end{minipage} & 
        \hspace{.3in}
        \raisebox{-.55in}{
            \includegraphics[width=1.1in]{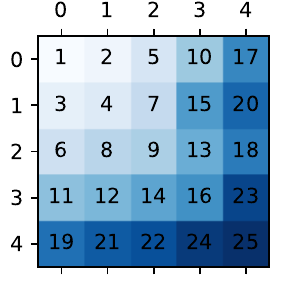}
        }
    \end{tabular}
    \caption{\textbf{Left:} maze prompt up to \texttt{<PATH\_START>}. \textbf{Right:} relative ordering of the cells in the vocabulary. Note that the top-left square of size $n \times n$ can be described using only the first $n^2$ tokens in the vocabulary.}
    \label{fig:output-tokenized}
\end{figure}

\hypertarget{benchmarks}{%
\section{Benchmarks of Generation Speed}\label{benchmarks}}

We provide approximate benchmarks for relative generation time across various algorithms, parameter choices, maze sizes, and dataset sizes.

\begin{table}[H]
  \centering
  \begin{tabular}{|l|c||r|r|r|r|}
    \hline
                          \multicolumn{2}{|c||}{Method \& Parameters}\Tstrut                       & \multicolumn{4}{c|}{Average time per maze (ms)} \\
    \hline 
    Generation algorithm  & Generation parameters & all sizes &  \begin{tabular}{@{}c@{}}small \\ ($g \leq 10$)\end{tabular} &  \begin{tabular}{@{}c@{}}medium \\ ($10 < g \leq 32$)\end{tabular} &  \begin{tabular}{@{}c@{}}large\\ ($g > 32$)\end{tabular}\\
    \hhline{|==||=|=|=|=|}
                 gen\_dfs\Tstrut  &   accessible\_cells=20 &     2.4 &           2.4 &            2.6 &           2.4 \\
                 gen\_dfs   &        do\_forks=False &     3.0 &           2.4 &            3.7 &           3.8 \\
                 gen\_dfs   &    max\_tree\_depth=0.5 &     4.5 &           2.2 &            4.9 &          11.6 \\
                 gen\_dfs   &           --          &    31.1 &           2.8 &           28.0 &         136.5 \\
      gen\_dfs\_percolation &                 p=0.1 &    53.9 &           3.6 &           42.5 &         252.9 \\
      gen\_dfs\_percolation &                 p=0.4 &    58.8 &           3.7 &           44.7 &         280.2 \\
         gen\_percolation   &           --          &    59.1 &           3.3 &           43.6 &         285.2 \\
              gen\_wilson   &           --          &   767.9 &          10.1 &          212.9 &        4530.4 \\
    \hhline{|==||=|=|=|=|}
    \multicolumn{2}{|c||}{\textbf{median (all runs)}}\Tstrut  & 10.8 &          6.0 &           44.4 &         367.7 \\
    \multicolumn{2}{|c||}{\textbf{mean (all runs)}}  & 490.0 &         11.7 &          187.2 &        2769.6 \\

    \hline
    \end{tabular}
    \vspace{1em}
    \caption{Average time to generate a single maze, averaged across multiple runs and dataset size. All benchmarks were run with parallelization disabled on a Intel i9-8950HK CPU.}
\end{table}

\begin{figure}[H]
    \centering
    \makebox[\linewidth][c]{
        \begin{subfigure}[b]{.453\textwidth}
            \centering
            \includegraphics[width=.95\textwidth]{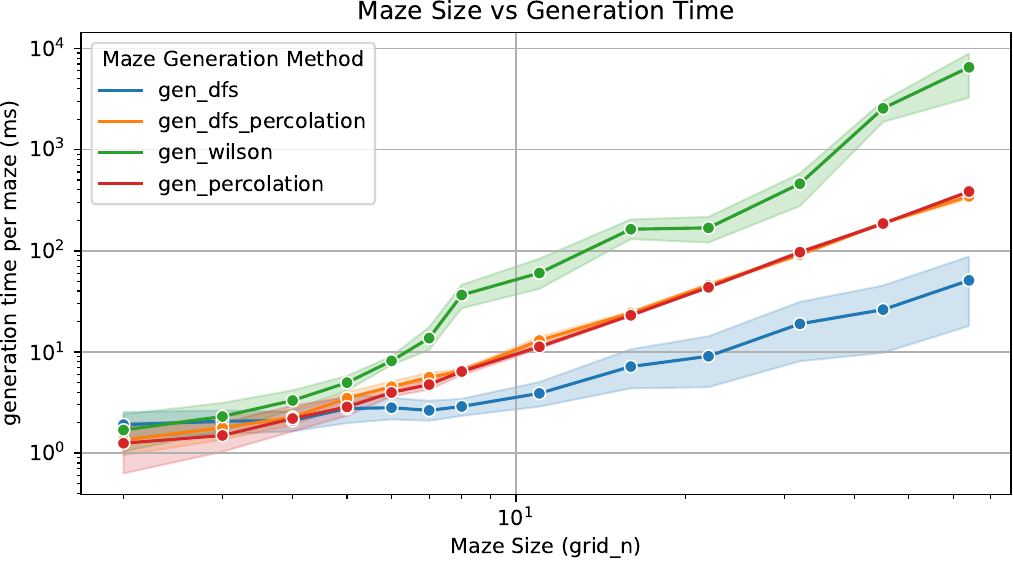}
        \end{subfigure}
        \begin{subfigure}[b]{.547\textwidth}
            \centering
            \includegraphics[width=.95\textwidth, trim={0 -0.79cm 0 0}]{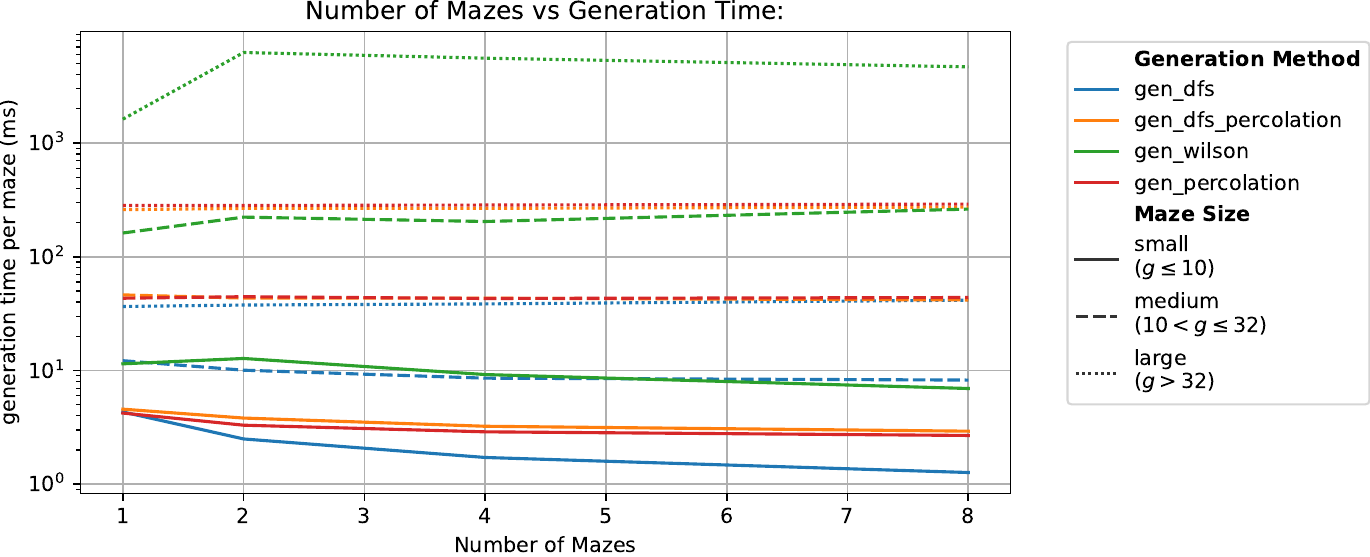}
        \end{subfigure}
    }
    \caption{Plots of maze generation time. Generation time scales exponentially with maze size for all algorithms (left). Generation time does not depend on the number of mazes being generated, and there is minimal overhead to initializing the generation process for a small dataset (right). Wilson's algorithm is notably less efficient than others and has high variance. Note that for both plots, values are averaged across all parameter sets for that algorithm, and parallelization is disabled.}
\end{figure}

\hypertarget{implementation}{%
\section{Implementation}\label{implementation}}

We refer to our GitHub repository \cite{maze-dataset-github} for documentation and up-to-date implementation details.

This package utilizes a simple, efficient representation of mazes. Using an adjacency list to represent mazes would lead to a poor lookup time of whether any given connection exists, whilst using a dense adjacency matrix would waste memory by failing to exploit the structure (e.g., only 4 of the diagonals would be filled in).
Instead, we describe mazes with the following simple representation: for a $d$-dimensional lattice with $r$ rows and $c$ columns, we initialize a boolean array $A = \{0, 1\}^{d \times r \times c}$, which we refer to in the code as a \texttt{connection\_list}. The value at $A[0,i,j]$ determines whether a downward connection exists from node $[i,j]$ to $[i+1, j]$. Likewise, the value at $A[1,i,j]$ determines whether a rightwards connection to $[i, j+1]$ exists. Thus, we avoid duplication of data about the existence of connections, at the cost of requiring additional care with indexing when looking for a connection upwards or to the left. Note that this setup allows for a periodic lattice.\footnote{
    That is, rather than a sub-graph of \(\mathbb{Z}^2\), we are working on the lattice \( \mathbb{Z}/r\mathbb{Z} \times \mathbb{Z}/c\mathbb{Z} \). This is achieved by using modular arithmetic for indexing. Specifically, when considering connections from a node at position \([i, j]\), the downward connection leads to the node at position \([ (i+1) \mod r, j]\), and the rightward connection leads to the node at position \([i, (j+1) \mod c]\). However, although our data structure supports this in principle, our algorithms for solving and visualizing the mazes do not. In practice, the last elements of $A$ are always set to $0$ to remove the possibility of periodic connections.
}

To produce solutions to mazes, two points are selected uniformly at random without replacement from the connected component of the maze, and the $A^*$ algorithm \cite{A_star} is applied to find the shortest path between them.

Parallelization is implemented via the \texttt{multiprocessing} module in the Python standard library, and parallel generation can be controlled via keyword arguments to the \texttt{MazeDataset.from\_config()} function.

\hypertarget{relation-work}{%
\subsection{Relation to Existing Works}\label{relation-work}}

As mentioned in the introduction, a multitude of public and open-source software packages exist for generating mazes~\cite{easy_to_hard, gh_Ehsan_2022, gh_Nemeth_2019}. However, our package provides more flexibility and efficiency in the following ways:
\begin{itemize}
    \item For rigorous investigations of the response of a model to various distributional shifts, preserving metadata about the generation algorithm with the dataset itself is essential. To this end, our package efficiently stores the dataset along with its metadata in a single human-readable file~\cite{zanj}. This metadata is loaded when the dataset is retrieved from disk and reduces the complexity of discerning the parameters under which a dataset was created.
    \item Prior works provide maze datasets in only a rasterized format, which is not suitable for training autoregressive text-based transformer models. As discussed in Section~\ref{output-formats}, our package provides these different formats natively.
    \item Our package provides a selection of maze generation algorithms, which all write to a single unified format. All output formats are reversible, and operate to and from this unified format.
\end{itemize}

As mentioned in Section~\ref{training}, we also include the \texttt{RasterizedMazeDataset} class in our codebase, which can exactly mimic the outputs provided in \texttt{easy-to-hard-data}~\cite{easy_to_hard}. Our \inlinecode{as\_ascii()} method provides a format similar to that used in~\cite{eval-gpt-visual}. The text format provided by \inlinecode{as\_tokens()} is similar to that of \cite{eval-LLM-graphs}, but provides a custom tokenization scheme.

\hypertarget{limitations}{%
\subsection{Limitations of \texttt{maze-dataset}}\label{limitations}}

For simplicity, the package primarily supports mazes that are sub-graphs of a 2-dimensional rectangular lattice. Some support for higher-dimensional lattices is present, but not all output formats are adapted for higher dimensional mazes.
As mentioned in Section~\ref{implementation}, our codebase does not fully support utilizing the periodic structure allowed by the data structure representing the maze.
Since the use of $A^*$ described in Section~\ref{implementation} does not have a preference between two paths of equal length, solutions to mazes which are not acyclic may not always be unique.

\hypertarget{conclusion}{%
\section{Conclusion}\label{conclusion}}

The \href{https://github.com/understanding-search/maze-dataset}{\inlinecode{maze-dataset}} library~\cite{maze-dataset-github} introduced in this paper provides a flexible and extensible toolkit for generating, processing, and analyzing maze datasets. By supporting various procedural generation algorithms and conversion utilities, it enables the creation of mazes with customizable properties to suit diverse research needs. 
Planned improvements to the \inlinecode{maze-dataset} include adding more generation algorithms (such as Prim's algorithm \cite{jarnik-prim, prim, dijkstra-prim} and Kruskal's algorithm \cite{kruskal}, among others \cite{mazegen_analysis}), adding the ability to augment a maze with an adjacency list to add ``shortcuts'' to the maze, and resolving certain limitations detailed in Section~\ref{limitations}.
Future work will make extensive use of this library to study interpretability and out-of-distribution generalization in autoregressive transformers~\cite{maze-transformer-github}, recurrent convolutional neural networks~\cite{deepthinking}, and implicit networks~\cite{mckenzie2023faster}.

\hypertarget{acknowledgements}{%
\section{Acknowledgements}\label{acknowledgements}}
First and foremost, the authors would like to thank each other for the good times had in developing this library and the subsequent research which was carried out. We are also indebted to AI Safety Camp and AI Safety Support for supporting this project and bringing many of the authors together.
This work was partially funded by National Science Foundation award DMS-2309810.
We thank the Mines Optimization and Deep Learning group (MODL) for fruitful discussions.

\newpage

\printbibliography

\newpage

\hypertarget{appendix}{%
\section{Appendix: Examples of Generated Mazes}\label{appendix}}

\begin{figure}[ht]

\centering
\caption{Results for Wilson's algorithm}

    \includegraphics[width=0.8\textwidth]{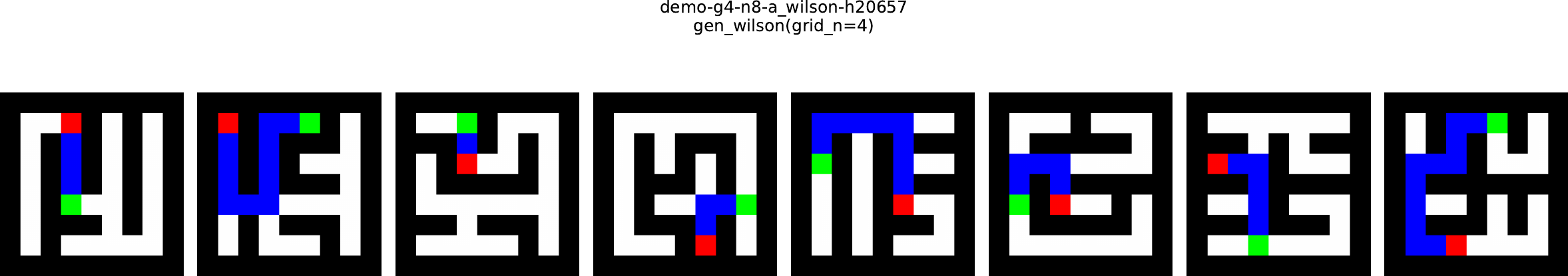} \\[1em]
    \includegraphics[width=0.8\textwidth]{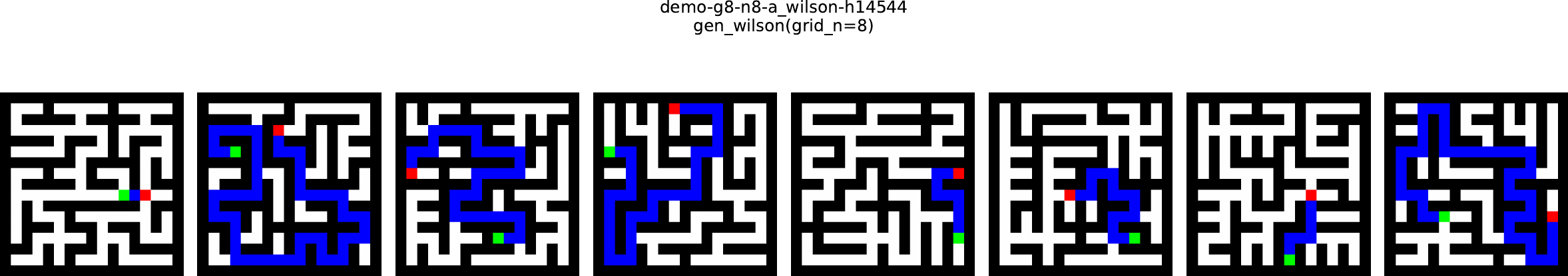} \\[1em]
    \includegraphics[width=0.8\textwidth]{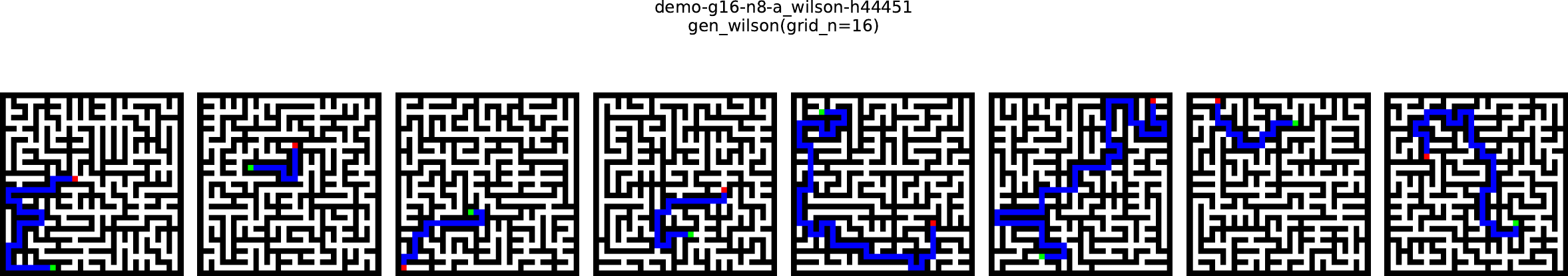} \\[1em]
    \includegraphics[width=0.8\textwidth]{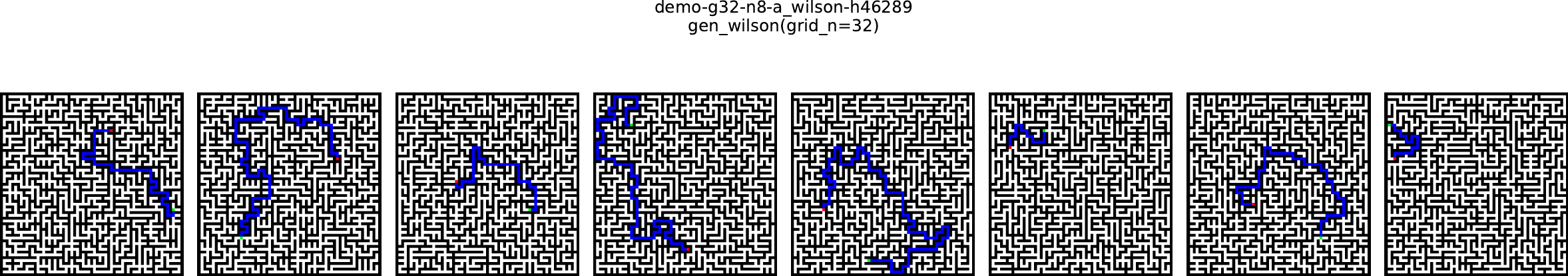} \\[1em]
\end{figure}

\begin{figure}[ht]

\centering
\caption{Results for randomized depth first search, with various parameters}

\begin{tabular}{c c}
    \begin{minipage}{0.49\textwidth}
        \centering \vspace{.5em}
        Standard RDFS: \\[.5em]
        \includegraphics[width=0.9\textwidth]{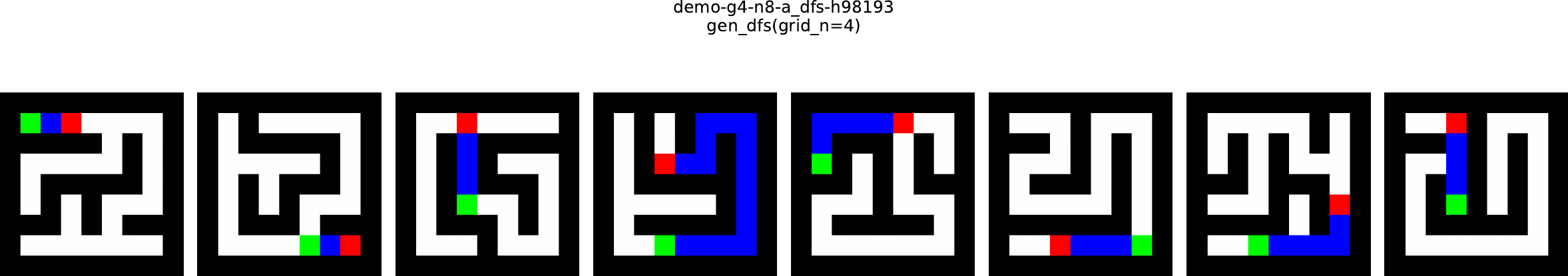} \\[1em]
        \includegraphics[width=0.9\textwidth]{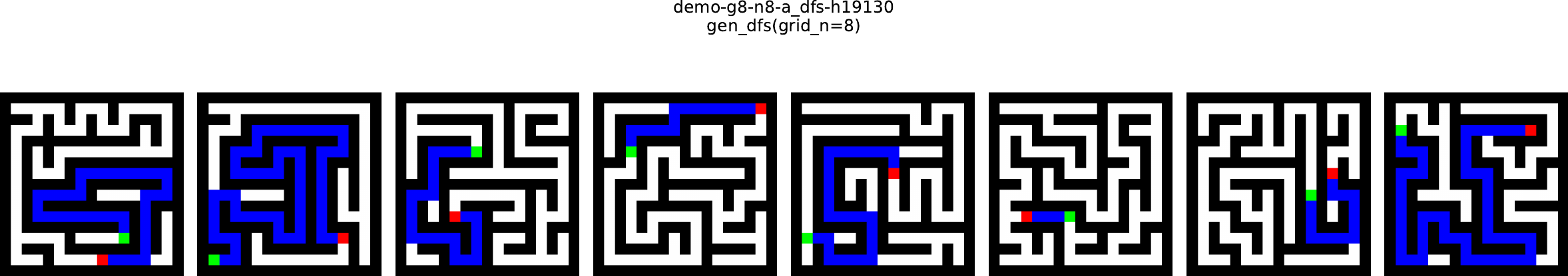} \\[1em]
        \includegraphics[width=0.9\textwidth]{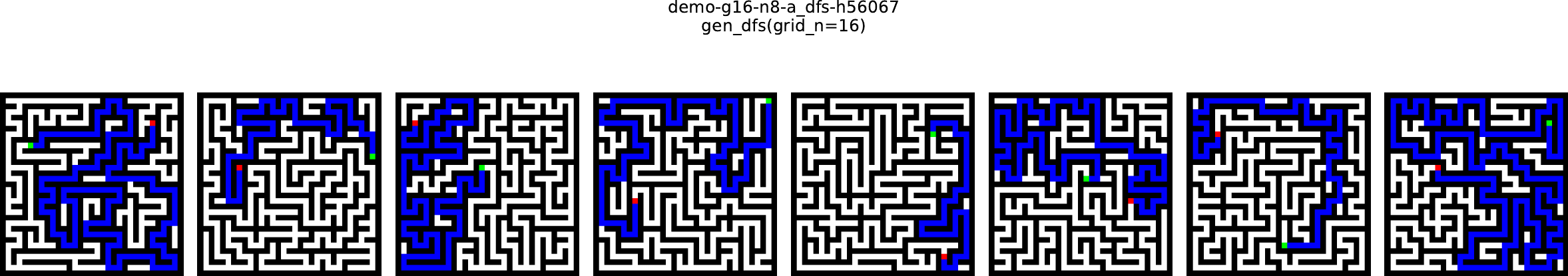} \\[1em]
        \includegraphics[width=0.9\textwidth]{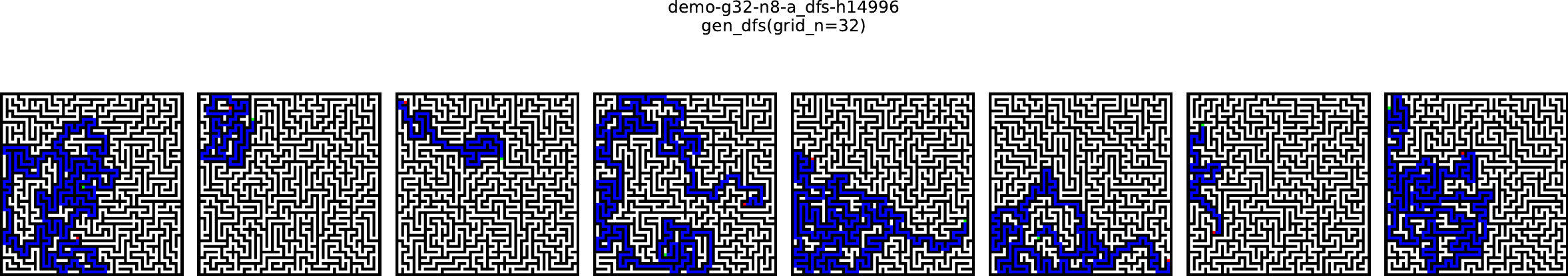}        
    \end{minipage}
    &
    \begin{minipage}{0.49\textwidth}
        \centering \vspace{.5em}
        RDFS without forks: \\[.5em]
        \includegraphics[width=0.9\textwidth]{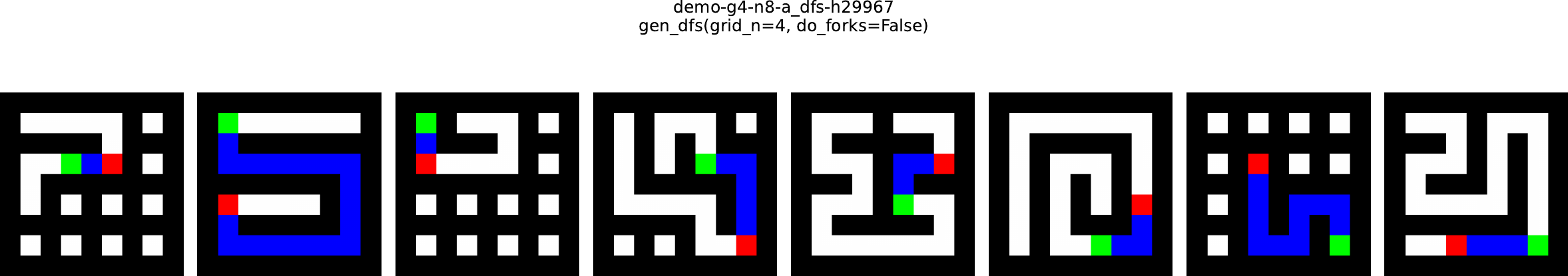} \\[1em]
        \includegraphics[width=0.9\textwidth]{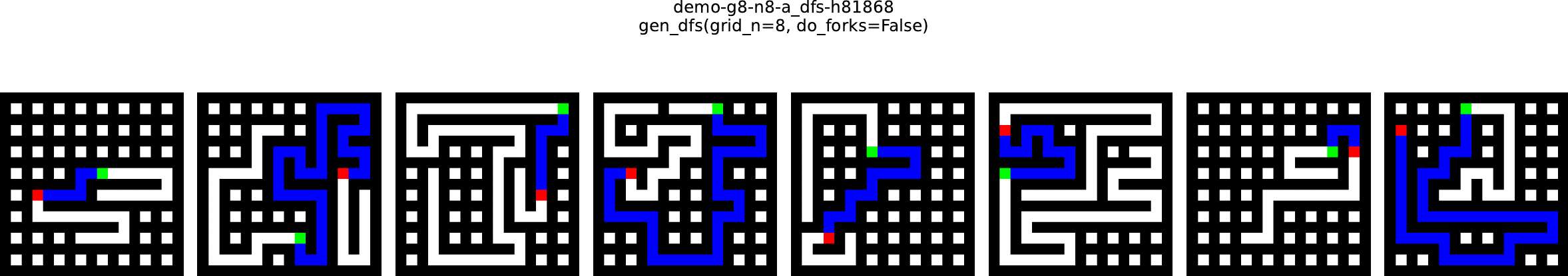} \\[1em]
        \includegraphics[width=0.9\textwidth]{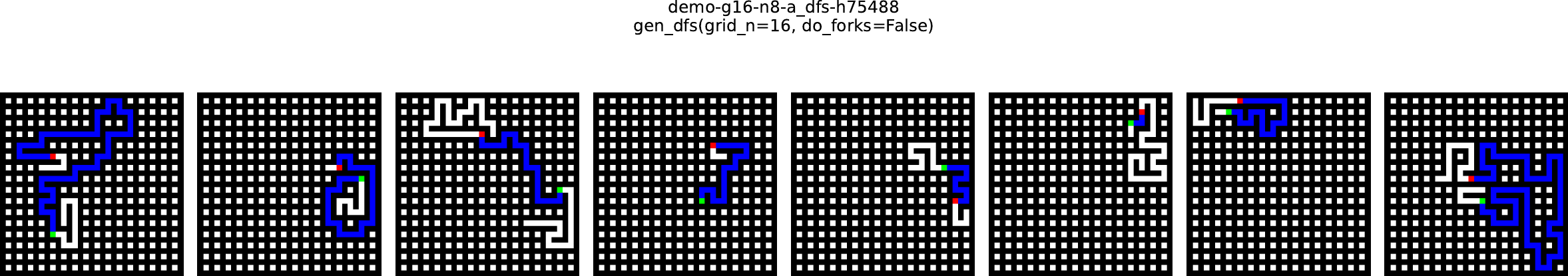} \\[1em]
        \includegraphics[width=0.9\textwidth]{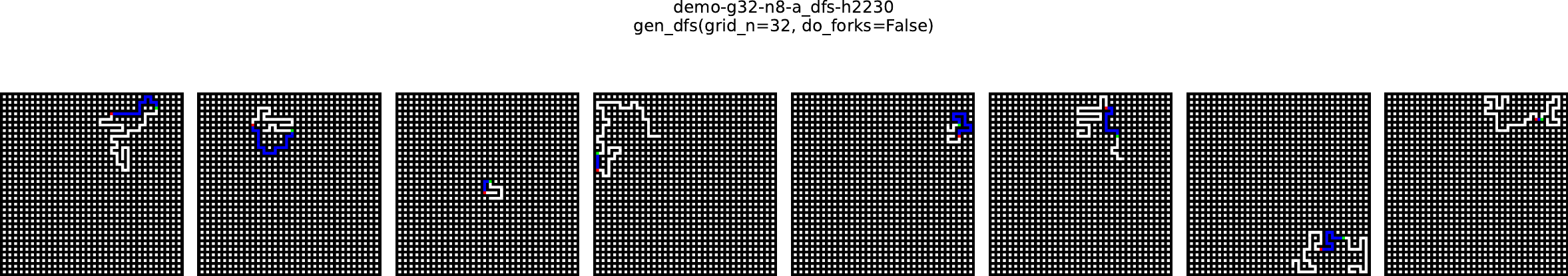} \\[1em]    
    \end{minipage}
    \\

    \begin{minipage}{0.49\textwidth}
        \centering \vspace{1em}
        RDFS with percolation, $p=0.1$: \\[.5em]
        \includegraphics[width=0.9\textwidth]{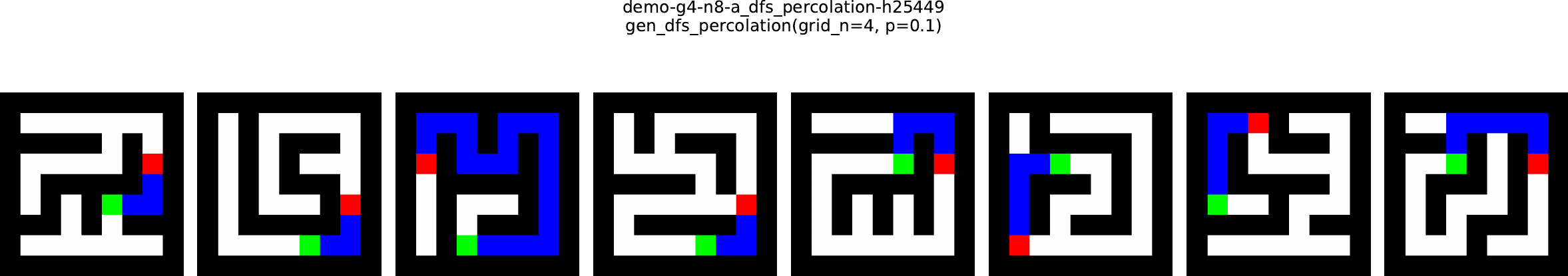} \\[1em]
        \includegraphics[width=0.9\textwidth]{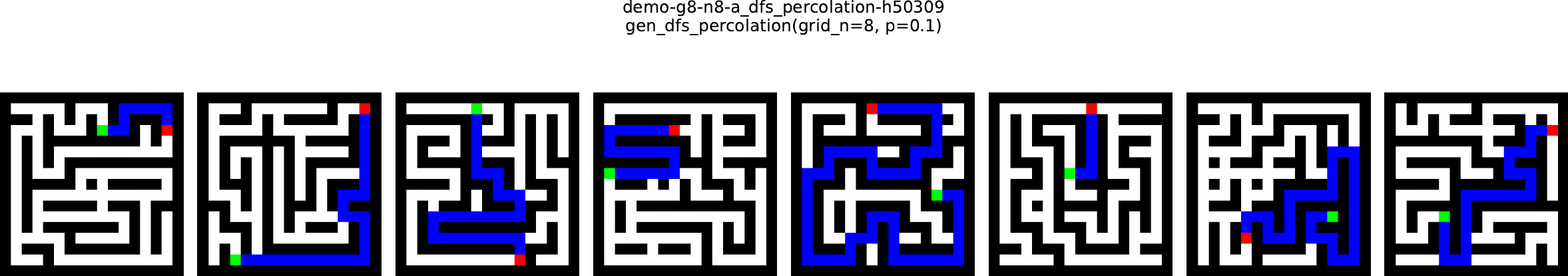} \\[1em]
        \includegraphics[width=0.9\textwidth]{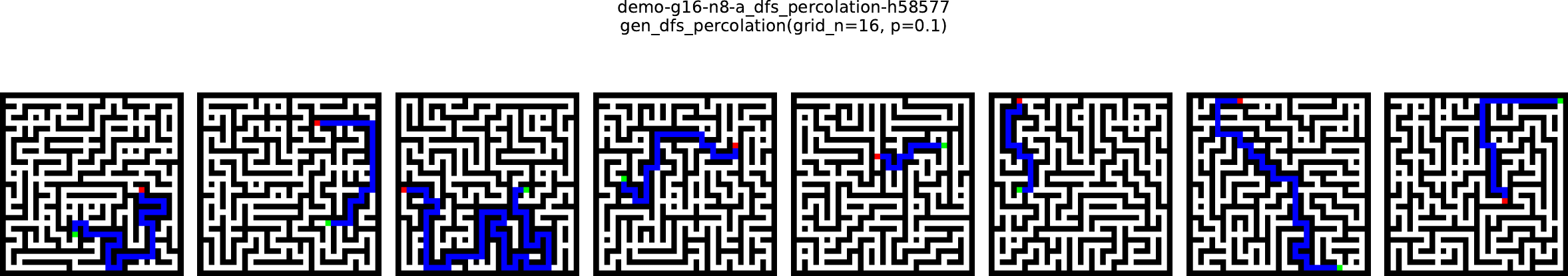} \\[1em]
        \includegraphics[width=0.9\textwidth]{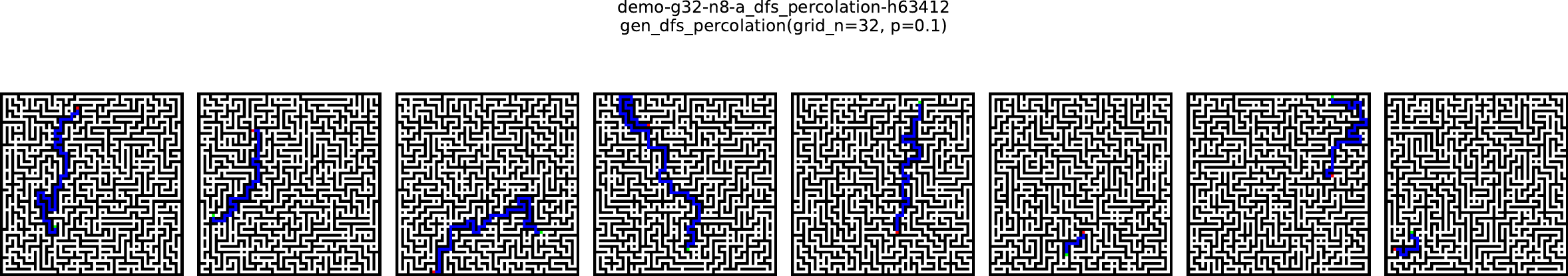} \\[1em]
    \end{minipage}
    &
    \begin{minipage}{0.49\textwidth}
        \centering \vspace{1em}
        RDFS with percolation, $p=0.4$: \\[.5em]
        \includegraphics[width=0.9\textwidth]{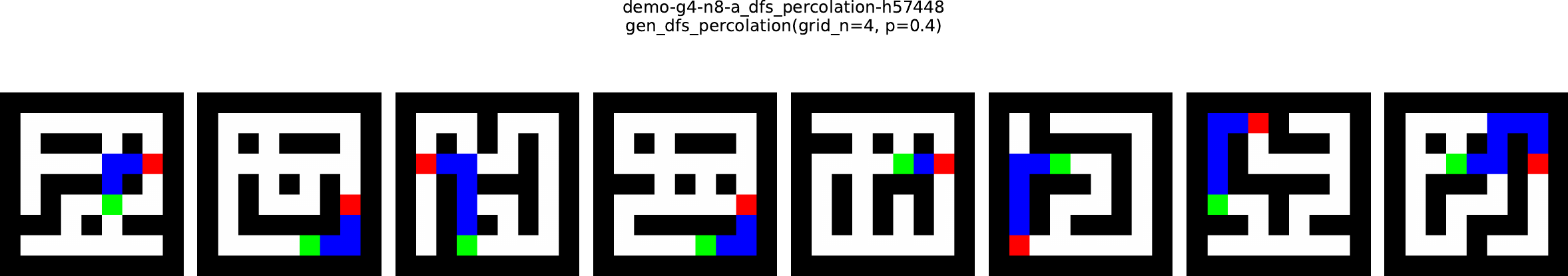} \\[1em]
        \includegraphics[width=0.9\textwidth]{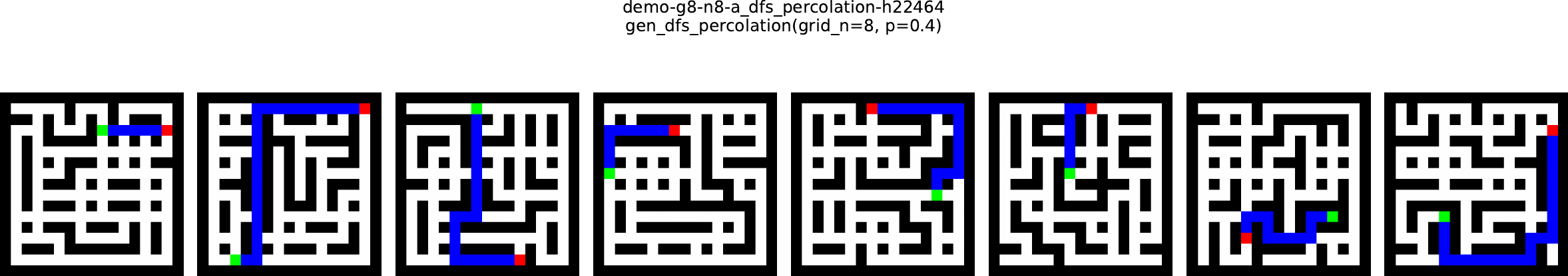} \\[1em]
        \includegraphics[width=0.9\textwidth]{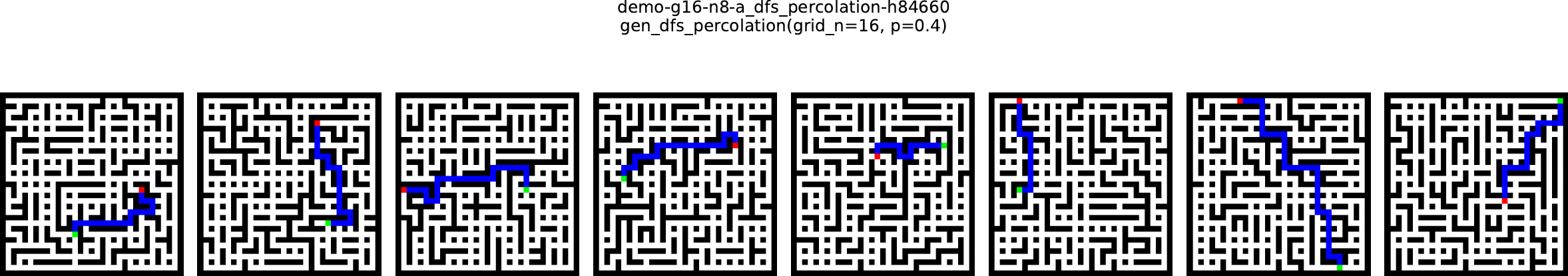} \\[1em]
        \includegraphics[width=0.9\textwidth]{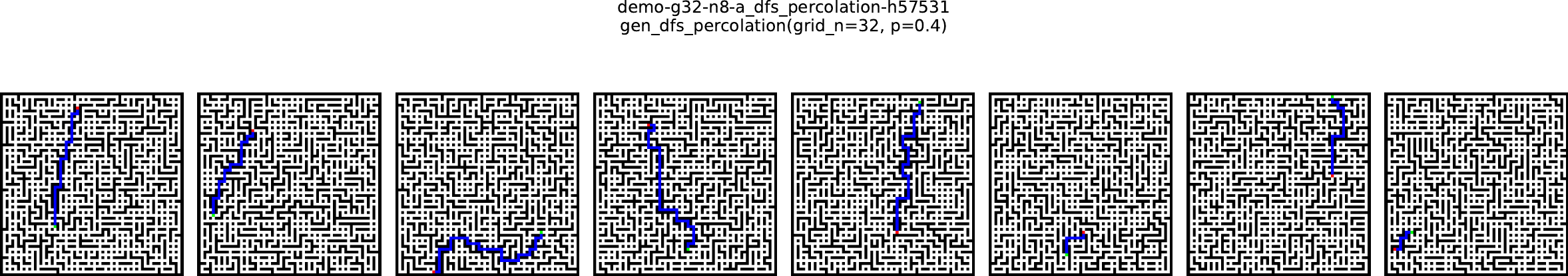} \\[1em]
    \end{minipage}
    \\

    \begin{minipage}{0.49\textwidth}
        \centering \vspace{1em}
        RDFS with maximum tree depth constrained: \\[.5em]
        \includegraphics[width=0.9\textwidth]{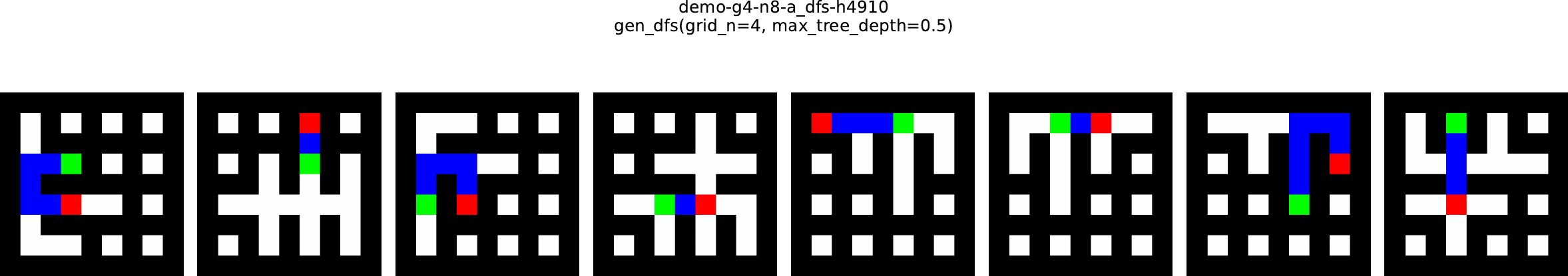} \\[1em]
        \includegraphics[width=0.9\textwidth]{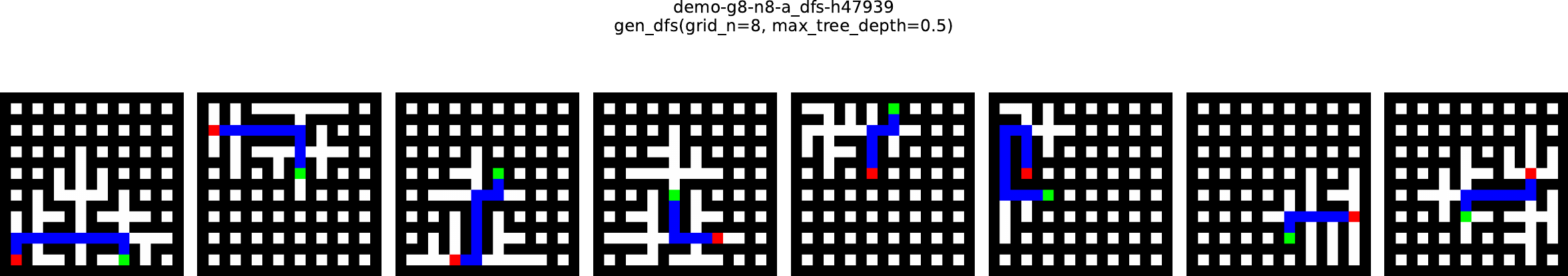} \\[1em]
        \includegraphics[width=0.9\textwidth]{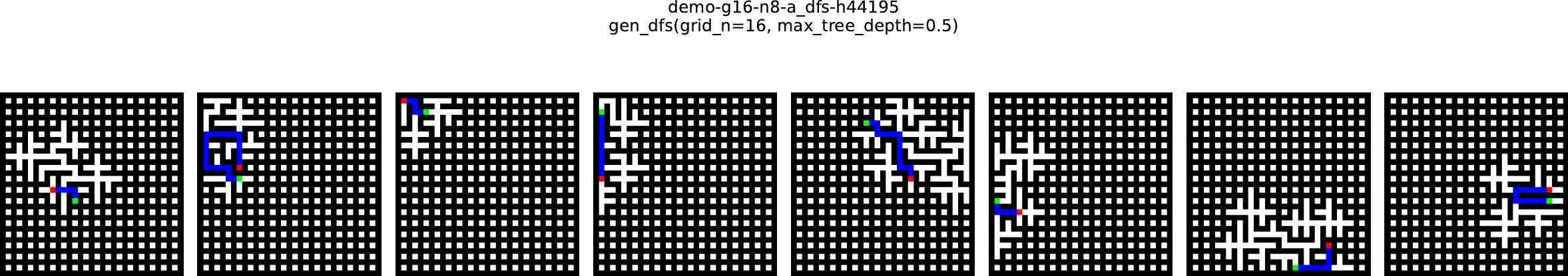} \\[1em]
        \includegraphics[width=0.9\textwidth]{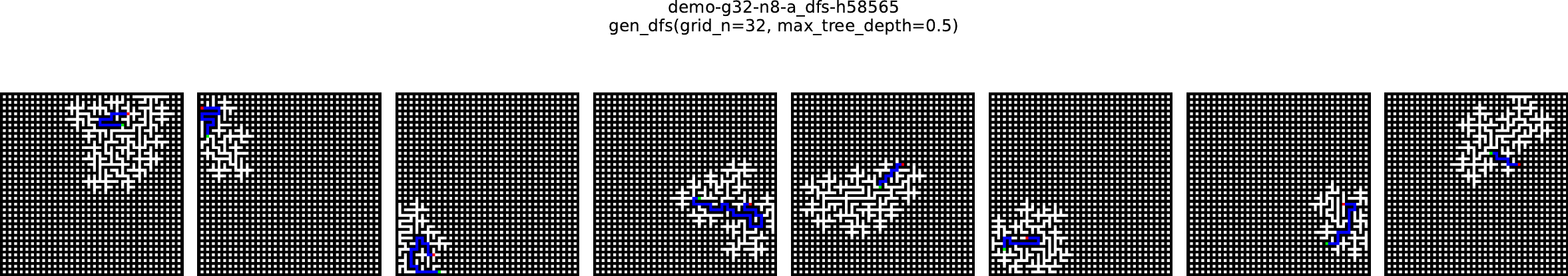} \\[1em]
    \end{minipage}
    &
    \begin{minipage}{0.49\textwidth}
        \centering \vspace{1em}
        RDFS with number of acessible cells constrained: \\[.5em]
        \includegraphics[width=0.9\textwidth]{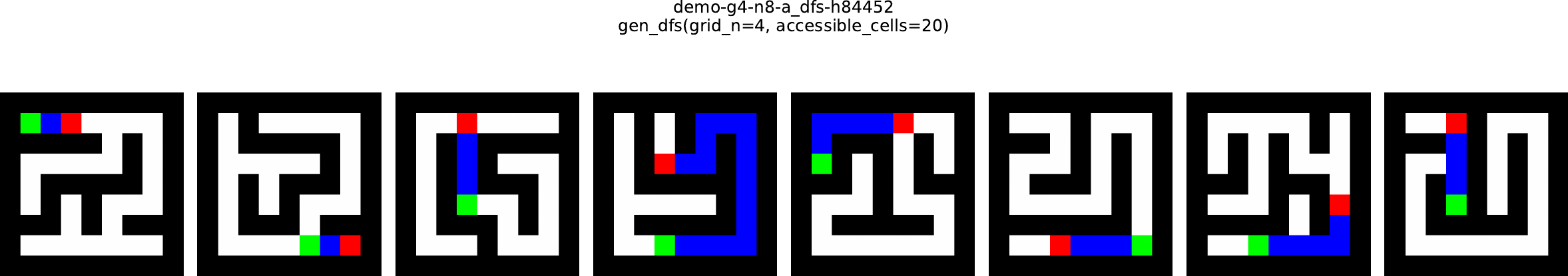} \\[1em]
        \includegraphics[width=0.9\textwidth]{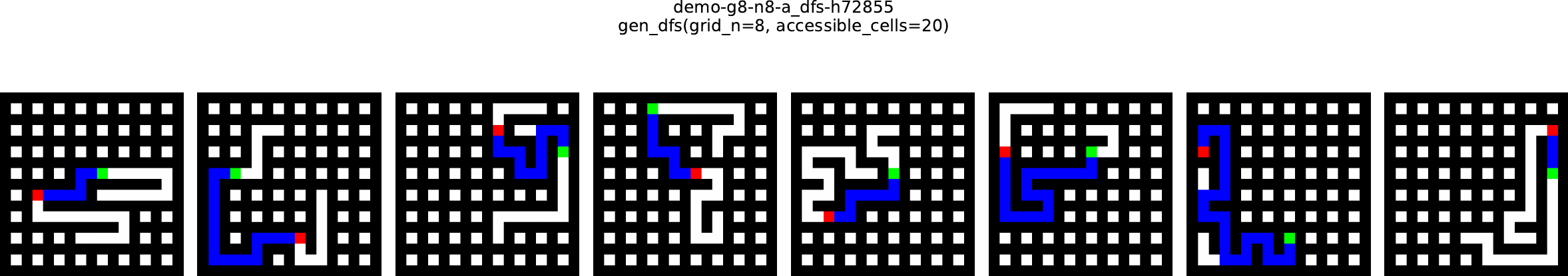} \\[1em]
        \includegraphics[width=0.9\textwidth]{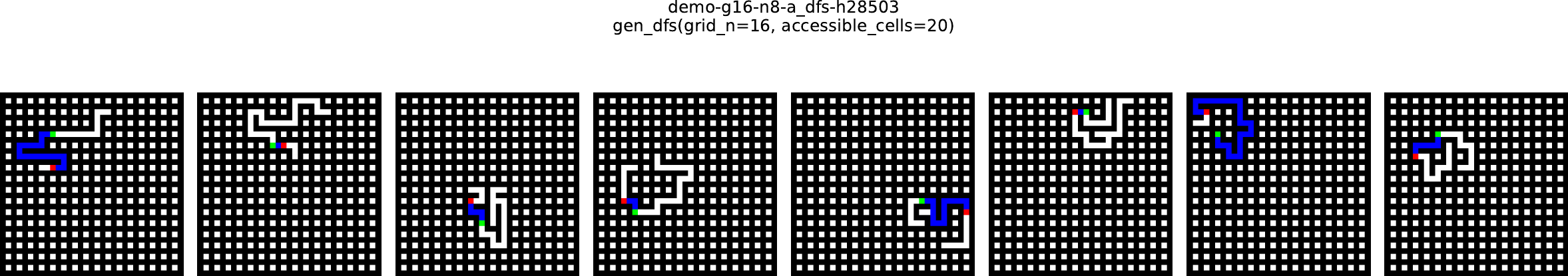} \\[1em]
        \includegraphics[width=0.9\textwidth]{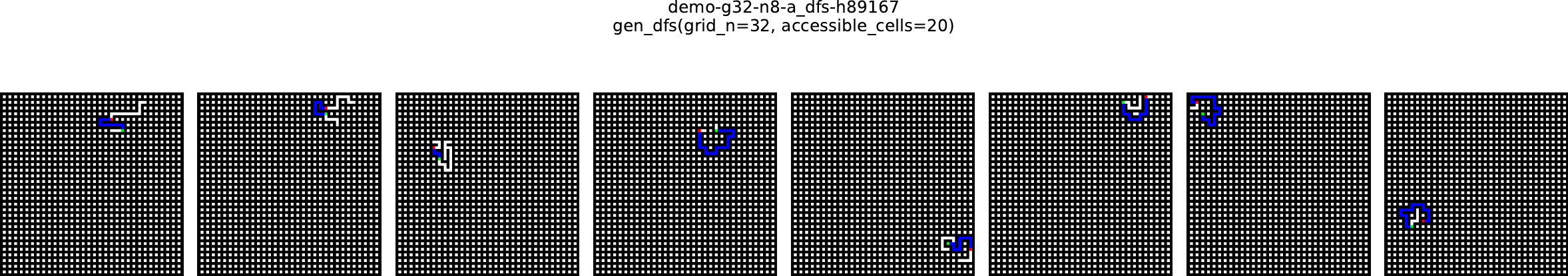} \\[1em]
    \end{minipage}

\end{tabular}

\end{figure}

\end{document}